\documentclass{article}
\usepackage{amsmath}
\usepackage{amssymb}
\usepackage{amsthm}
\usepackage{array,makecell}
\usepackage{booktabs}
\usepackage{caption}
\usepackage{float}
\usepackage{gensymb}
\usepackage{graphicx}
\usepackage{lineno}
\usepackage{mathtools}
\usepackage{microtype}
\usepackage{multirow,bigdelim}
\usepackage{soul}
\usepackage{subfigure}
\usepackage{xspace}
\usepackage[textsize=tiny]{todonotes}
\usepackage{hyperref}
\usepackage[capitalize,noabbrev]{cleveref}

\theoremstyle{plain}

\theoremstyle{definition}

\theoremstyle{remark}

\def\arxivversion{}

\newcommand{\fd}{{MAV3D}\xspace}
\newcommand{\nerf}{{NeRF}\xspace}
\newcommand{\mav}{{Make-A-Video}\xspace}

\newcommand\Tau{\mathrm{T}}

\newcommand{\myparagraph}[1]{\noindent\textbf{#1}}

\ifdefined\arxivversion 
\newcommand{\github}{{\href{https://make-a-video3d.github.io}{make-a-video3d.github.io}}\xspace}
\else 
\newcommand{\github}{{\href{https://gen4D.github.io}{gen4D.github.io}}\xspace}
\fi

\ifdefined\arxivversion 
\usepackage[accepted]{icml2023}
\else 
\usepackage{icml2023}
\icmltitlerunning{Submission and Formatting Instructions for ICML 2023} 
\fi

\begin{document}
\twocolumn[{%
\icmltitle{Text-To-4D Dynamic Scene Generation}




\icmlsetsymbol{equal}{*}

\begin{icmlauthorlist}
\icmlauthor{Uriel Singer}{equal}
\icmlauthor{Shelly Sheynin}{equal}
\icmlauthor{Adam Polyak}{equal}
\icmlauthor{Oron Ashual}{}
\icmlauthor{Iurii Makarov}{}
\icmlauthor{Filippos Kokkinos}{}
\icmlauthor{Naman Goyal}{}
\icmlauthor{Andrea Vedaldi}{}
\icmlauthor{Devi Parikh}{}
\icmlauthor{Justin Johnson}{}
\icmlauthor{Yaniv Taigman}{}
\end{icmlauthorlist}

\ifdefined\arxivversion
\else
\icmlaffiliation{equal}{random, Location, Country}
\icmlcorrespondingauthor{Firstname1 Lastname1}{first1.last1@xxx.edu}
\icmlcorrespondingauthor{Firstname2 Lastname2}{first2.last2@www.uk}
\fi

\icmlkeywords{Machine Learning, ICML}

\vskip 0.3in
\begin{center}
  \vspace{-2mm}
  \includegraphics[height=0.35\textwidth]{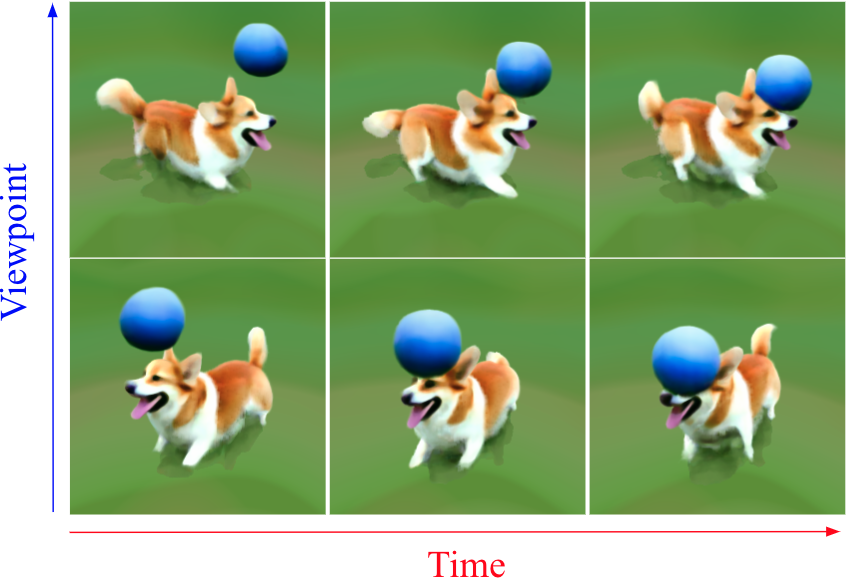}
  \includegraphics[height=0.35\textwidth]{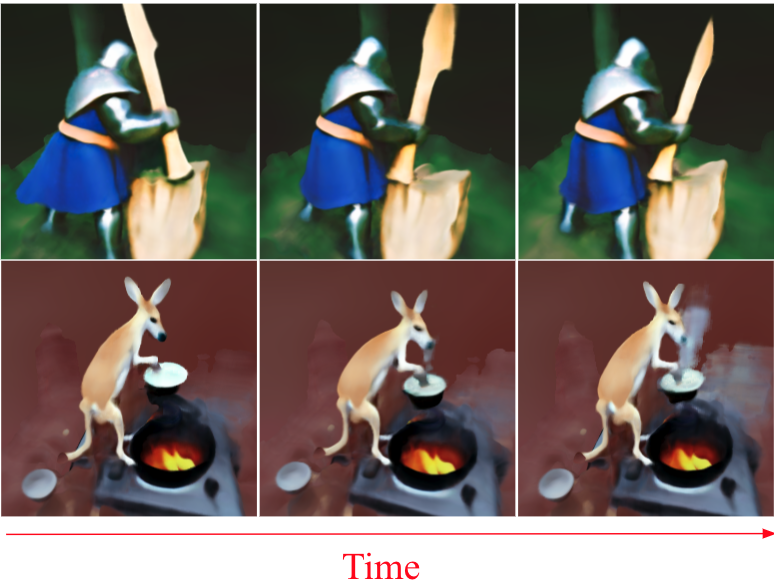}
  \vspace{-2mm}
  \captionof{figure}{Samples generated by \fd along the temporal and viewpoint dimensions. Left: ``A corgi playing with a ball''. Right top:``A knight chopping wood''. Right bottom: ``A kangaroo cooking a meal''. \\}
  \label{fig:teaser}
\end{center}
}
]


\printAffiliationsAndNotice{\icmlEqualContribution} 



\begin{abstract}
We present \fd (\textbf{M}ake-\textbf{A}-\textbf{V}ideo\textbf{3D}), a method for generating three-dimensional dynamic scenes from text descriptions. Our approach uses a 4D dynamic Neural Radiance Field (NeRF), which is optimized for scene appearance, density, and motion consistency by querying a Text-to-Video (T2V) diffusion-based model.
The dynamic video output generated from the provided text can be viewed from any camera location and angle, and can be composited into any 3D environment.
\fd does not require any 3D or 4D data 
and the T2V model is trained only on Text-Image pairs and unlabeled videos.
We demonstrate the effectiveness of our approach using comprehensive quantitative and qualitative experiments and show an improvement over previously established internal baselines.
To the best of our knowledge, our method is the first to generate 3D dynamic scenes given a text description. Generated samples can be viewed at {\github}.

\end{abstract}

\section{Introduction}
\label{intro}


Generative models have seen tremendous recent progress,
and can now generate realistic images from natural language prompts~\cite{ramesh2022hierarchical,gafni2022make,rombach2022high,saharia2022photorealistic,yu2022scaling,sheynin2022knn}.
This success has been extended beyond 2D images
both \emph{temporally} to synthesize videos~\cite{singer2022make,ho2022imagen} and \emph{spatially} to produce 3D shapes~\cite{poole2022dreamfusion,lin2022magic3d,nichol2022point}.
However, these two categories of generative models have been studied in isolation to date.

In this paper we combine the benefits of video and 3D generative models and propose a novel system for \emph{text-to-4D} (3D$+$time) generation.
Our method, named \fd (\textbf{M}ake-\textbf{A}-\textbf{V}ideo\textbf{3D}),
takes as input a natural-language description and outputs a dynamic 3D scene representation which can be rendered from arbitrary viewpoints.
Such a method could be used to generate animated 3D assets for video games, visual effects, or augmented and virtual reality.


%
Differently from image and video generation where one can train on large quantities of captioned data, there is no readily available collection of 4D models, with or without textual annotations.
One approach might be to start from a pre-trained 2D video generator~\cite{singer2022make} and distill a 4D reconstruction from generated videos.
%
Still, reconstructing the shape of deformable objects from video is a very challenging, widely known as Non-Rigid Structure from Motion (NRSfM).
The task becomes simpler if one is given \emph{multiple simultaneous viewpoints} of the object.
While multi-camera setups are rare for real data, our insight is that existing video generators \emph{implicitly model arbitrary viewpoints for generated scenes.}
We can thus use a video generator as a `statistical' multi-camera setup to reconstruct the geometry and photometry of the deformable object.
Our \fd algorithm does so by optimizing a dynamic Neural Radiance Field (NeRF) jointly with decoding the input text into a video, sampling random viewpoints around the object.
%

Naively optimizing dynamic NeRF using video generators does not produce satisfying results and there are several significant challenges that must be overcome toward this goal.
First, we need an effective \emph{representation} for dynamic 3D scenes that is efficient and learnable end-to-end.
Second, we need a source of \emph{supervision} since there are no large-scale datasets of (text, 4D) pairs from which to learn.
Third, we need to scale the \emph{resolution} of the outputs in both space and time which is both memory- and compute-intensive due to the 4D output domain.



For our \emph{representation}, we build on recent advances in neural radiance fields (NeRFs)~\cite{mildenhall2021nerf}.
We combine insights from work on efficient (static) NeRFs~\cite{sun2022direct,mueller2022instant} and dynamic NeRFs~\cite{cao2023hexplane}, and represent a 4D scene as a set of six multiresolution feature planes.




To \emph{supervise} this representation without paired (text, 4D) data, we propose a multi-stage training pipeline for dynamic scene rendering and demonstrate the importance of each component in achieving high-quality results. One key observation is that directly optimizing a dynamic scene using Score Distillation Sampling (SDS)~\cite{poole2022dreamfusion} using Text-to-Video (T2V) model leads to visual artifacts and sub-optimal convergence. Therefore, we first utilize a Text-to-Image (T2I)~\cite{singer2022make} model to fit a static 3D scene to a text prompt and subsequently augment our 3D scene model with dynamics. Additionally, we introduce a new temporal-aware SDS loss and motion regularizers that prove to be crucial for realistic and challenging motion.

We scale to higher \emph{resolution} outputs with an additional phase of temporal-aware super-resolution fine-tuning.
We use SDS from the super-resolution module of the T2V model to obtain high-resolution gradient information to supervise our 3D scene model, increasing its visual fidelity and allowing us to sample higher-resolution outputs during inference.

Our main contributions are:
\begin{itemize}
    \setlength{\itemsep}{0pt}
    \setlength{\parskip}{0pt}
    \item We introduce \fd, an effective method that utilizes T2V model and dynamic NeRFs in order to integrate world knowledge into 3D temporal representations.
    \item We propose a multi-stage static-to-dynamic optimization scheme that gradually incorporates gradient information from static, temporal, and super-resolution models, to enhance the 4D scene representation.
    \item We conduct a comprehensive set of experiments, including ablation studies, using both quantitative and qualitative metrics to reveal the technical decisions made during the development of our method. 
\end{itemize}
\vspace{-3mm}

\begin{figure*}[ht]
    \centering
    \setlength{\tabcolsep}{1.5pt}
    \scalebox{0.9}{
    \begin{tabular}{c c }
        \vspace{-0.1cm}\includegraphics[width=0.49\textwidth]{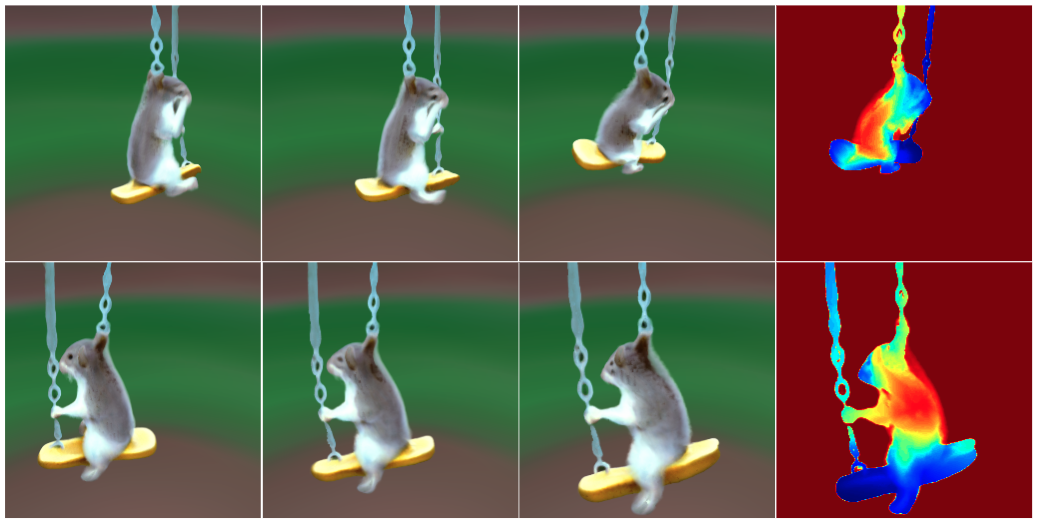} &
        \hspace{0.1cm}
        \includegraphics[width=0.49\textwidth]{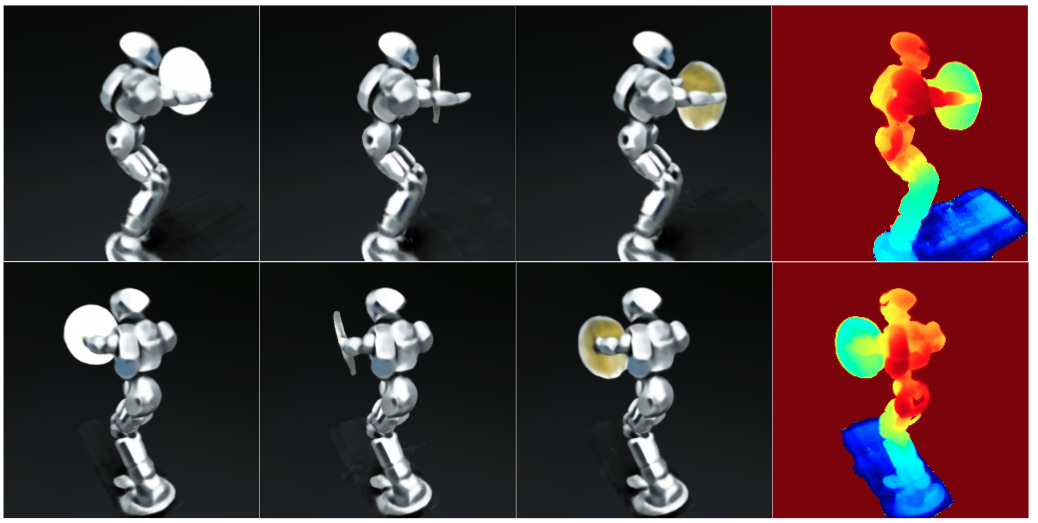} \\
        \vspace{0.2cm} a squirrel playing on a swing set & silver humanoid robot flipping a coin \\
        \vspace{-0.1cm} \includegraphics[width=0.49\textwidth]{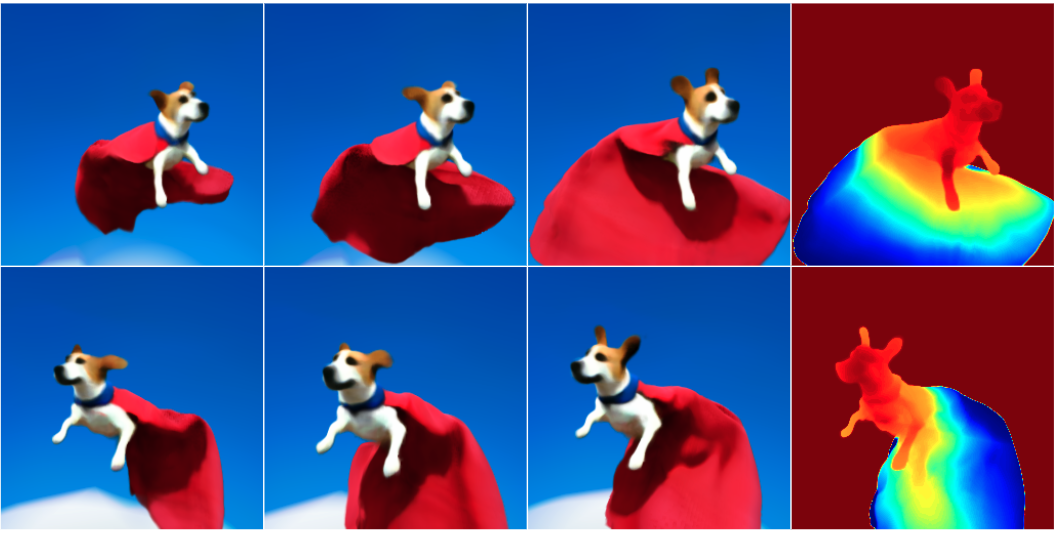} &
        \hspace{0.1cm}
        \includegraphics[width=0.49\textwidth]{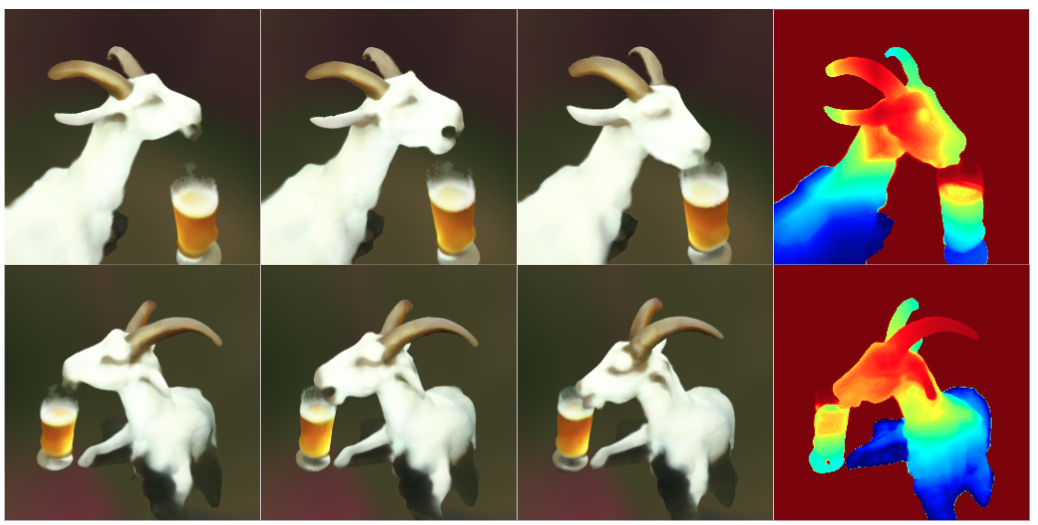} \\
        \vspace{0.2cm} superhero dog with red cape flying through the sky & a goat drinking beer \\
        \vspace{-0.1cm} \includegraphics[width=0.49\textwidth]{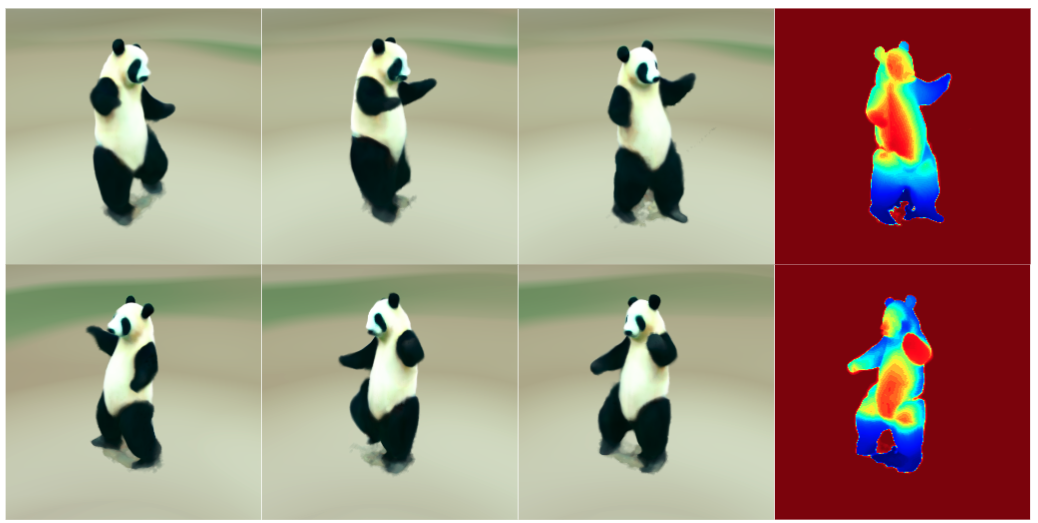} &
        \hspace{0.1cm}
        \includegraphics[width=0.49\textwidth]{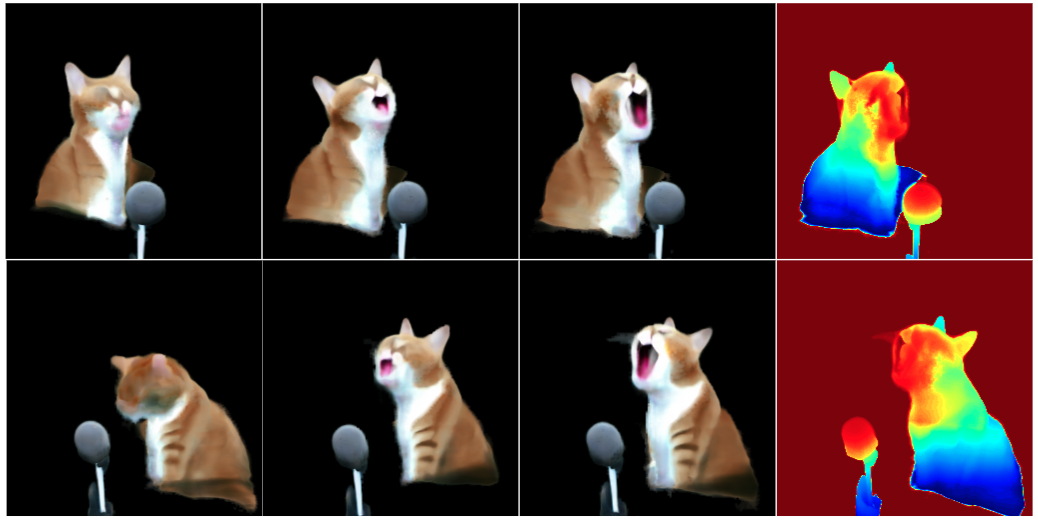} \\
        a panda dancing & a cat singing
    \end{tabular}
    }
    \vspace{-0.2cm}
    \caption{Samples generated by \fd. The rows represent variations in time, and the columns represent variations in viewpoint. The last column shows the depth image of its adjacent column. 
    } 
    \vspace{-4mm}
    \label{fig:scene_sketching_comparisons}
\end{figure*}

\section{Related work}
\label{prev}


\myparagraph{Neural rendering.}
Our 3D scene representation builds upon recent advances in neural rendering.
Neural radiance fields (NeRFs) \cite{mildenhall2021nerf}
represent a 3D scene with a neural network that inputs scene coordinates, and form images with volume rendering.
Recent work has improved efficiency by incorporating 3D data structures such as voxel grids~\cite{sun2022direct} that may be sparse~\cite{fridovich2022plenoxels} or multiresolution~\cite{takikawa2021nglod}, and which can be further accelerated via tensor factorization~\cite{chen2022tensorf} or hashing~\cite{mueller2022instant}.

\myparagraph{Dynamic neural rendering.}
We aim to generate \emph{dynamic} scenes which can be viewed from any angle.
This relates to classic work on free-viewpoint video which use videos of a moving scene to synthesize novel views~\cite{carranza2003free,smolic20063d,collet2015high}.
Recent approaches make NeRFs dynamic
by conditioning the network on both space and time~\cite{martin2021nerf,li2022neural} and may incorporate additional supervision from depth~\cite{xian2021space} or scene flow~\cite{li2021neural,du2021neural,gao2021dynamic}.
Another category of approaches learn a time-varying \emph{deformation} of 3D points into a static \emph{canonical} scene~\cite{pumarola2021d,park2021nerfies,tretschk2021nonrigid,park2021hypernerf}.
Some approaches accelerate NeRFs on dynamic scenes using explicit voxel grids~\cite{fang2022fast} or tensor factorization~\cite{cao2023hexplane,fridovich2023kplanes}.

\myparagraph{Text to 3D.}
The idea of generating 3D scenes from text dates back decades~\cite{adorni1983natural}; early efforts parsed geometric relations from text and built scenes from a library of known objects~\cite{coyne2001wordseye,chang2014learning}.
Some approaches train neural networks end-to-end on paired datasets of text and shape~\cite{chen2018text2shape,nichol2022point} but this approach is difficult to scale due to the paucity of paired data.
Other approaches instead generate 3D shapes from text without paired data using a pretrained CLIP~\cite{radford2021learning} model~\cite{jetchev2021clipmatrix,sanghi2022clip,wang2022clip,jain2022dreamfields} or a text-to-image diffusion model~\cite{poole2022dreamfusion,lin2022magic3d}.
We use a similar strategy, but generate 4D rather than 3D content using a text-to-video model.

\myparagraph{Diffusion-based generative models.} 
Recent improvements in diffusion models~\cite{dhariwal2021diffusion} have led to highly advanced image synthesis~\cite{ramesh2021zero, esser2021taming, rombach2022high, gafni2022make, nichol2021glide, ramesh2022hierarchical, saharia2022photorealistic} and the creation of generative models for other forms of media, such as video\cite{singer2022make, ho2022imagen, phenaki}. Our video generator is based on Make-A-Video~(MAV)~\cite{singer2022make}, which expands upon a Text-To-Image (T2I) model by training on unlabeled videos.



\begin{figure*}[ht!]
    \centering
    \includegraphics[
    width=0.95\textwidth]{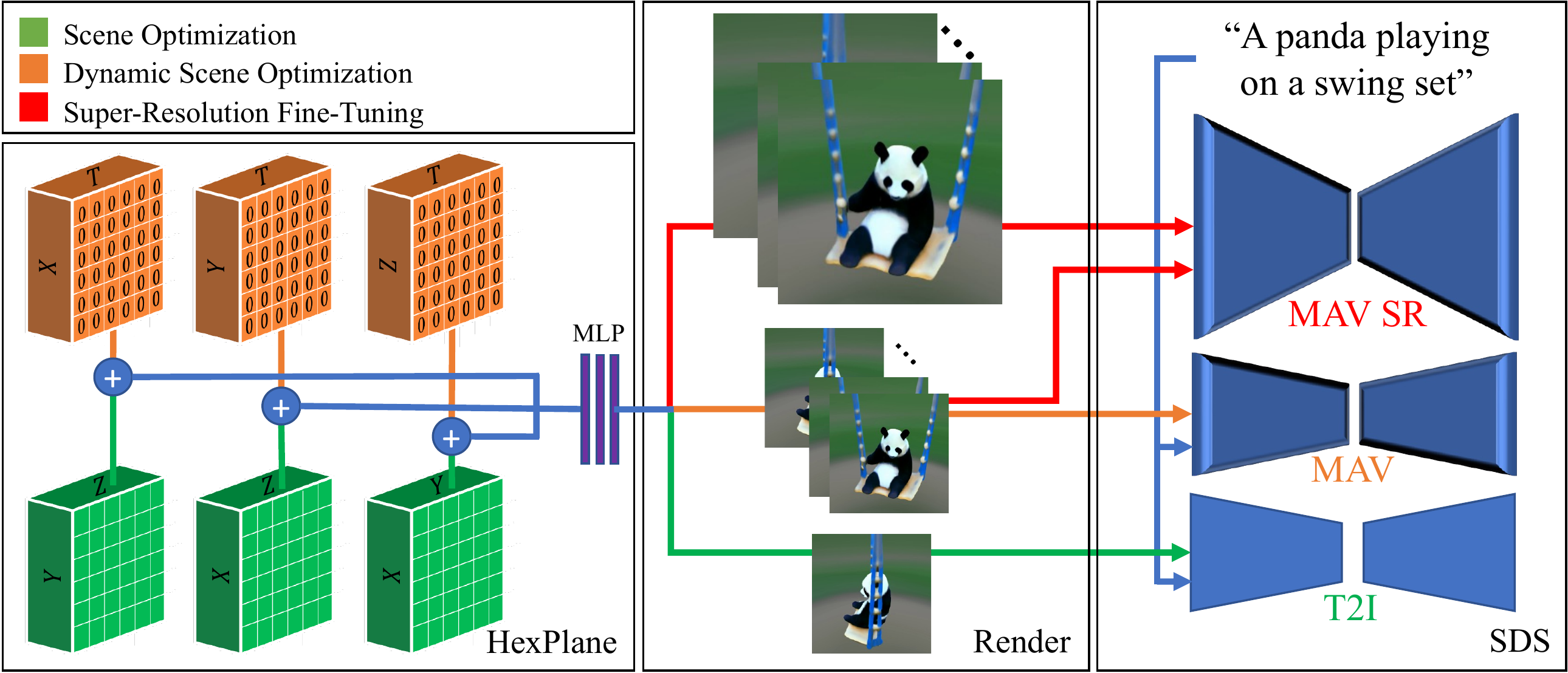}
    \caption{\textbf{Full pipeline of \fd.} First, we leverage only the three pure-spatial planes (colored green), render a single image, and calculate the SDS loss using the T2I model. In the second stage, we add the additional three planes (colored orange) which are initialized to zeros for smooth transition, render a full video and calculate the SDS-T loss using the T2V model. In the third stage (SRFT), we additionally render high-resolution video which is passed as input to the super-resolution component, with the low-resolution as condition.}
    \label{fig:mav3d_pipe}
    \vspace{-3mm}
\end{figure*}

\newcommand{\fps}{\mathrm{fps}}

\section{Method}%
\label{sec:method}

Our goal is to develop a method that produces a dynamic 3D scene representation from a natural-language description.
This is a challenging task since we have neither (text, 3D) pairs nor dynamic 3D scene data for training.
Instead, we rely on a pretrained \emph{text-to-video} (T2V) diffusion model~\cite{singer2022make} as a scene prior, which has learned to model realistic appearance and motion of scenes by training on large-scale image, text, and video data.

At a high level, given a text prompt $p$ we fit a 4D scene representation $f_\theta(x, y, z, t)$ that models the appearance of a scene matching the prompt at arbitrary points in spacetime.
Without paired training data, we cannot directly supervise the outputs from $f_\theta$; however given a sequence of camera poses
$\{C_t\}_{t=1}^T$
we can \emph{render} a sequence of images
$I_t = \mathcal{R}(f_\theta, t, C_t)$
from $f_\theta$ and stack them to form a video $V$.
Then, we can pass the text prompt $p$ and the video $V$ to a frozen, pretrained T2V diffusion model which scores the video's realism and alignment to the prompt; we can then use \emph{Score Distillation Sampling} (SDS)~\cite{poole2022dreamfusion} to compute an update direction for the scene parameters $\theta$.

The above pipeline can be seen as an extension of DreamFusion~\cite{poole2022dreamfusion}, adding a temporal dimension to the scene model and supervising with a a T2V model rather than a \emph{text-to-image} (T2I) model.
However, we found that high-quality text-to-4D generation requires additional significant innovations.
First, we use a new 4D scene representation (\cref{sec:representation}) which allows flexible modeling of scene motion.
Second, we enhance video quality and improve model convergence with a multi-stage \emph{static-to-dynamic} optimization scheme (\cref{sec:static-to-dynamic}) that utilizes several \emph{motion regularizers} to encourage realistic motion.
Third, we improve the \emph{resolution} of the model using \emph{super-resolution fine-tuning} (SRFT) (\cref{sec:srft}). See Fig.~\ref{fig:mav3d_pipe} for an illustration of the method. 


\subsection{4D Scene Representation}%
\label{sec:representation}

Following recent advances in neural rendering~\cite{mildenhall2021nerf}, we represent a dynamic 3D scene \emph{implicitly}.
Given a timestep $t$ and camera position, for each image pixel we cast a ray through the camera plane into the scene and sample a set of points
$\{\mu_i\}_{i=1}^N$
along the ray; for each point we compute a \emph{volume density} $\tau_i\geq0$ and color $c_i$,
and the color for the ray is computed via volume rendering.
Similar to NeRF~\cite{mildenhall2021nerf}, we output the color $c_i$ from an MLP, but assume that the color is view-independent (Lambertian).
We explored generating albedo (scene color) and random light sources like DreamFusion~\cite{poole2022dreamfusion} but found that it significantly slows training without improving quality.
Our learnable scene model is thus a function $(\tau, c_i) = f_\theta(x, y, z, t)$ that outputs volume density and color for arbitrary points in spacetime.

We must then choose a suitable architecture for $f_\theta$.
DreamFusion~\cite{poole2022dreamfusion} adopts a variant of Mip-NeRF~\cite{barron2022mip} for recovering static 3D structure from images.
By analogy, we can adopt any architecture for recovering dynamic 3D structure from videos.
Such architectures are often designed to operate with as few as one input view, and thus include strong scene priors (\textit{e.g.}, temporal deformations from a static canonical scene~\cite{pumarola2021d,park2021nerfies}) that enable reconstruction from sparse views.

However, in our setting we need not learn from sparse views; SDS on a pretrained T2V model enables us to supervise the model along arbitrary view trajectories during training.
As such, rather than adopting an architecture which regularizes and restricts scene motion, we instead would like a high-capacity architecture that can flexibly model large scene motion.
We thus adopt HexPlane~\cite{cao2023hexplane}, a recently proposed representation for dynamic scenes.

\myparagraph{HexPlane}
represents a 4D scene with six \emph{planes} of feature vectors spanning all pairs of axes in $\{X, Y, Z, T\}$.
It computes an $(R_1+R_2+R_3)$-dimensional feature for a spacetime point $(x, y, z, t)$ via projection onto each plane:
{\footnotesize
\begin{equation}
  [P^{XYR_1}_{xy} + P^{ZTR_1}_{zt};
   P^{XZR_2}_{xz} + P^{YTR_2}_{yt};
   P^{YZR_3}_{yz} + P^{XTR_3}_{yz}]
   \label{eq:hexplane}
\end{equation}}%
Superscripts denote shapes ($P^{XYR_1}$ has shape $X{\times}Y{\times}R_1$, and is a plane of $R_1$-dim features spanning the $XY$ axes), subscripts denote sampling via bilinear interpolation, and ${;}$ is concatenation.
The resulting feature is passed to a small MLP which predicts volume density and color.

We further increase the capacity of the HexPlane model by representing each plane as a multi-resolution grid, similar to~\cite{mueller2022instant} with the hash function removed.
We also use a \emph{background model} simulating a large (static) sphere surrounding the (dynamic) foreground modeled by the HexPlane;
it is a small MLP receiving a sinusoidally-encoded ray direction and producing an RGB color.
We also keep a coarse voxel grid of occupancy probabilities updated periodically via EMA to accelerate sampling.
See Sec.~\ref{implement_details} of the supplementary for more details.


\subsection{Dynamic Scene Optimization}%
\label{sec:static-to-dynamic}



Armed with the HexPlane model $f_\theta$ for 4D scenes, we must \emph{supervise} it to match a textual prompt $p$.
We introduce temporal \emph{Score Distillation Sampling} (SDS-T) which is an extension of SDS~\cite{poole2022dreamfusion} for pretrained conditional video generator. We first describe the loss, and then its application in \fd. 


Recall that, given the camera trajectory $C$ and the scene parameters $\theta$, the rendering function $\mathcal{R}$ infers a parametric video $V_\theta$.
We assume that the pretrained conditional video generator is based on diffusion and thus defines a \emph{denoising function}
$
\hat{\epsilon}(V_{(\theta,\sigma,\epsilon)} \mid y, \sigma)
$
which takes as input a noised video
$
V_{(\theta,\sigma,\epsilon)}
=
\sqrt{1 - \sigma^2} V_\theta + \sigma \epsilon
$
(where $\epsilon\in\mathcal{N}(0,I)$ is normal noise),
the noise level
$
\sigma\in(0,1),
$
and additional conditioning information $y$ (textual prompt, video frame rate, etc.),
and predicts an estimate $\hat \epsilon$ of the noise $\epsilon$.

We compute an update direction for the scene parameters $\theta$ using SDS\@:
we add random noise $\epsilon$ to the current video $V_{\bar \theta}$ to obtain the noised version $V_{(\theta,\sigma,\epsilon)}$, apply the denoiser network to obtain the noise estimate $\hat \epsilon$.
The update direction for $\theta$ is then the (negative) gradient of the reconstruction loss
$
E_{\sigma,\epsilon}
[w(\sigma) \|\hat{\epsilon}(V_{(\textrm{sg}(\theta),\sigma,\epsilon)} \mid y, \sigma) - \epsilon \|^2]
$
averaged over $\epsilon$ and $\sigma$,
where $w(\sigma)$ is a weighing function and $\textrm{sg}(\cdot)$ is the stop-grad operator.%
\footnote{\cite{poole2022dreamfusion} use a more complex definition of $\mathcal{L}_\text{SDS}$; this simpler form gives the same gradient up to a scale factor.}
The resulting SDS gradient is then
{\footnotesize
\begin{equation}
\nabla_\theta \mathcal{L}_{SDS-T}
= 
E_{\sigma,\epsilon}
\left[
w(\sigma)
(
\hat{\epsilon}(V_{(\bar \theta,\sigma,\epsilon)} | y, \sigma)
- \epsilon
)
\dfrac{\partial V_\theta}{\partial \theta}
\right].
\label{eq:sds-t}
\end{equation}}%
We use the -T suffix to emphasise that, differently from~\cite{poole2022dreamfusion}, this version of SDS is applied to a video, i.e., a temporal image sequence.

\myparagraph{Static to dynamic.}
In practice we found that directly optimizing a HexPlane using SDS from a pretrained T2V model leads to visual artifacts and sub-optimal convergence (see Sec.~\ref{sec:ablation_sec}).
We therefore adopt a multi-stage \emph{static-to-dynamic} optimization scheme, first optimizing a \emph{static} 3D scene matching the text prompt, then extending it to 4D.

During the first phase of static optimization, we fix the three temporal planes of the HexPlane to zero
($P^{ZTR_1}$, $P^{YTR_2}$, and $P^{XTR_3}$ in Equation~\ref{eq:hexplane}); this is similar to the tri-plane representation used by~\cite{chan2022efficient}.
During each training iteration we sample a batch of 8 random camera poses, and render a $64{\times}64$ image from each view, applying view-dependent prompt engineering similar to DreamFusion.
We supervise the model using SDS on the frozen T2I model from~\cite{singer2022make} 
which has been pretrained on a large dataset of images and text.

During the second phase of dynamic optimization, we continue updating all planes of the HexPlane.
We render a batch of 8 $64{\times}64$ 16-frame videos from the model, and supervise it using SDS-T on the frozen T2V model from~\cite{singer2022make}. 
We employ several \emph{regularizers} during this phase to encourage high-quality 4D synthesis.

\myparagraph{Dynamic Camera.}
Most videos (including those on which our T2V model was trained) have apparent motion from two sources:
\emph{object motion} and \emph{camera motion}.
During training we simulate camera motion by rendering videos from randomly generated dynamic camera trajectories.

Training with dynamic cameras gives 4D scenes with more pronounced and realistic object motion (See Section~\ref{sec:ablation_sec}).
We hypothesize that, if trained with a static camera, the HexPlane tries to model object \emph{and} camera motion to close the domain gap with the T2V model, giving worse object motion.
Dynamic cameras also reduce the \emph{multi-face} problem common in DreamFusion:
the T2V model can judge a video showing faces on both sides of an object as unrealistic.

\myparagraph{FPS Sampling.}
Dynamic cameras randomize the \emph{spatial} perspective from which we render videos during training;
we also vary the \emph{temporal} extent of training videos via FPS sampling.
The T2V model accepts videos with $F{=}16$ frames, but also conditions on the \emph{frame-rate} of those videos.
For each training sample we randomly sample a frame-rate $\fps\sim\mathcal{U}[0, 1/F]$
and start time $t_0\sim\mathcal{U}[0, 1-F\cdot \fps]$,
where the 4D scene model assumes a temporal extent $t\in[0, 1]$.

\myparagraph{Gaussian Annealing.}
DreamFusion biases toward central scene content by adding Gaussian-distributed density to the output from the scene model before rendering;
we also use this bias during static optimization.
However in dynamic scenes, content should be allowed to move away from the origin;
we thus find it helpful to linearly increase the width of the Gaussian density bias during dynamic optimization.

\myparagraph{Total Variation Loss.}%
\label{sec:totalvarloss}
We encourage spacetime smoothness in our 4D representing by applying the following Total Variation (TV) regularizer (following~\cite{niemeyer2022regnerf}) to each of the six planes $P$ of the HexPlane:
{\footnotesize
\begin{equation}
  \mathcal{L}_{\textrm{TV}}(P;\beta)= \sum_{i,j}\left((P_{i,j+1}-P_{i,j})^2 + (P_{i+1,j}-P_{i,j})^2\right)^\frac{\beta}{2}
\end{equation}}%
The standard TV norm is obtained for $\beta=1$; however, we found that this resulted in noisy high-frequency artifacts in our case, similar to~\cite{mahendran15understanding}; we follow the latter and set $\beta=2$ to encourage smoothness.

\vspace{-2mm}
\subsection{Super-Resolution Fine-Tuning}%
\label{sec:srft}
During the static and dynamic scene optimization phases described above, our 4D scene representation is supervised via low-resolution $64\times64$ renderings;
we found that rendering higher-resolution videos from the learned model can lack detail and exhibit artifacts.
We overcome this problem with a final phase of \emph{super-resolution fine-tuning} (SRFT).

During SRFT we make use of the pretrained and frozen video super-resolution module $\textrm{SR}^t_l$ from~\cite{singer2022make}.
This diffusion-based model inputs a high-resolution noisy $256{\times}256$ video along with a clean $64{\times}64$ low-res video,
and predicts the noise of the high-resolution video.

We use $\textrm{SR}^t_l$ to improve high resolution renderings from our 4D scene model.
During each training iteration we sample a $256{\times}256$ video $V_{\uparrow}$ from our scene model and downsample it to a $64{\times}64$ video $V_{\downarrow}$.
These are used to compute an SDS gradient for $V_{\uparrow}$ using $\textrm{SR}^t_l$.

\begin{figure}[t!]
    \centering
    \setlength{\tabcolsep}{1.5pt}
    \scalebox{0.9}{
    \begin{tabular}{cc}
        \hspace{-0.2cm}\includegraphics[width=0.24\textwidth]{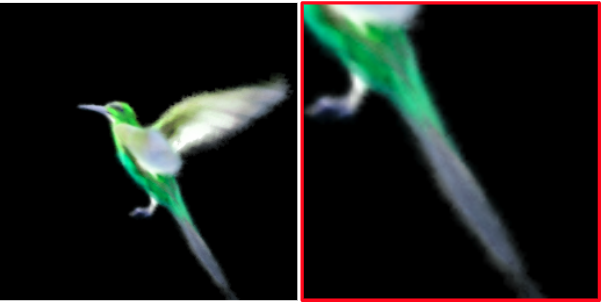} & \vspace{-0.1cm}\includegraphics[width=0.24\textwidth]{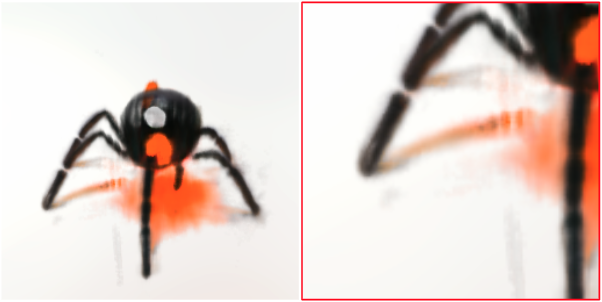} \\
        \hspace{-0.2cm}\includegraphics[width=0.24\textwidth]{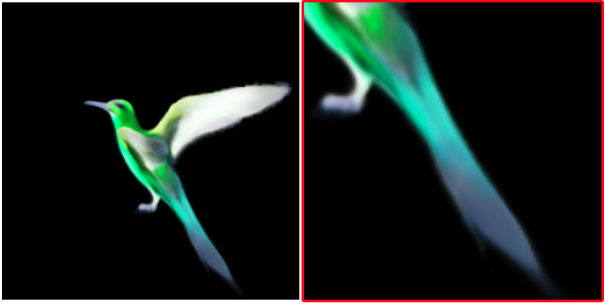} &
        \vspace{-0.1cm}\includegraphics[width=0.24\textwidth]{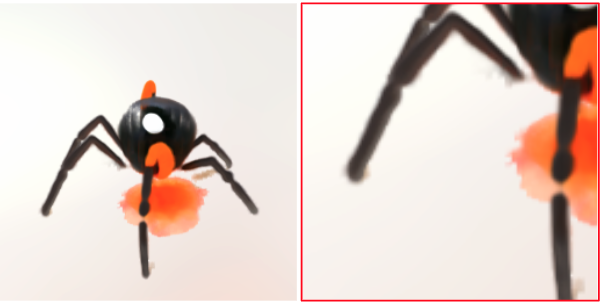} \\
        green hummingbird flying in & a dancing spider  \\
        \vspace{0.1cm}space and fluttering its wings & for halloween \\
        \hspace{-0.2cm}\includegraphics[width=0.24\textwidth]{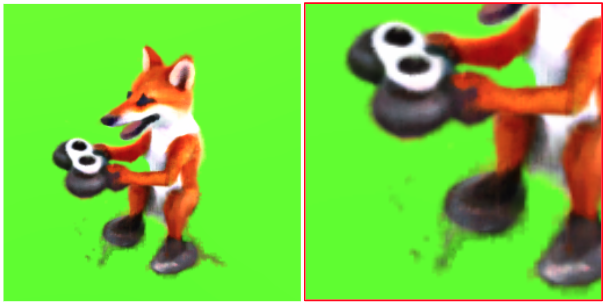} &
        \vspace{-0.1cm}\includegraphics[width=0.24\textwidth]{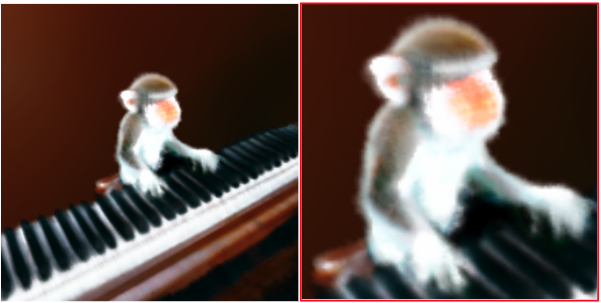} \\
        \hspace{-0.2cm}\includegraphics[width=0.24\textwidth]{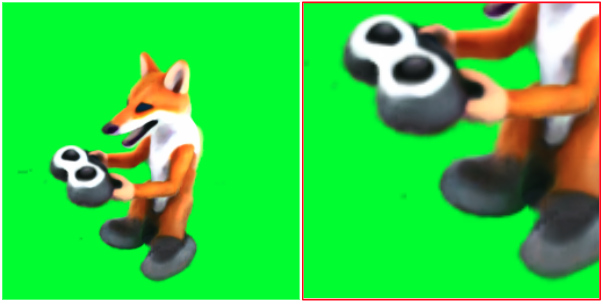} &
        \vspace{-0.1cm} \includegraphics[width=0.24\textwidth]{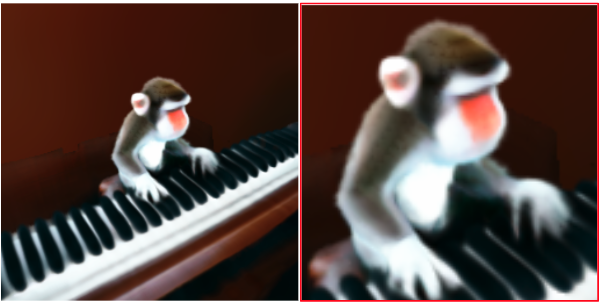} \\ 
        3D rendering of a & monkey learning  \\
        fox playing videogame & to play the piano \\
    \end{tabular}
    }
    \vspace{-2mm}
    \caption{An ablation on \textbf{temporal-aware super-resolution optimization}. \textbf{Top:} without super-resolution phase. \textbf{Bottom:} MAV3D results. The red square is a zoom-in area of the image. This stage enhances the quality of the rendered videos, resulting in
    high-resolution videos with finer details
    } 
    \vspace{-2mm}
    \label{fig:SR_ablat_fig}
\end{figure}

$\textrm{SR}^t_l$ does not condition on the text prompt $p$.
Fine-tuning via SDS on $\textrm{SR}^t_l$ alone thus encourages realistic high-resolution videos, but not alignment to $p$; this can cause the model to collapse to an empty scene.
During SRFT we therefore train jointly using SDS from $\textrm{SR}^t_l$ and SDS-T from Equation~\ref{eq:sds-t}.

\section{Experiments}

Our experiments evaluate the ability of \fd to generate dynamic scenes from text descriptions.
First, we evaluate the effectiveness of our approach on the Text-To-4D task.
To the best of our knowledge \fd is the first method to tackle this task, so we develop three alternative methods as baselines.
Second, we evaluate simplified versions of our model on the sub-tasks of T2V and Text-To-3D, and compare them to existing baselines in the literature.
Third, we conduct a comprehensive ablation study to justify our method's design. Fourth, we describe our procedure for converting dynamic NeRFs into dynamic meshes, and finally present an extension of our model to the Image-to-4D task.

\begin{figure}[t]
\centering
\includegraphics[width=0.48\textwidth]{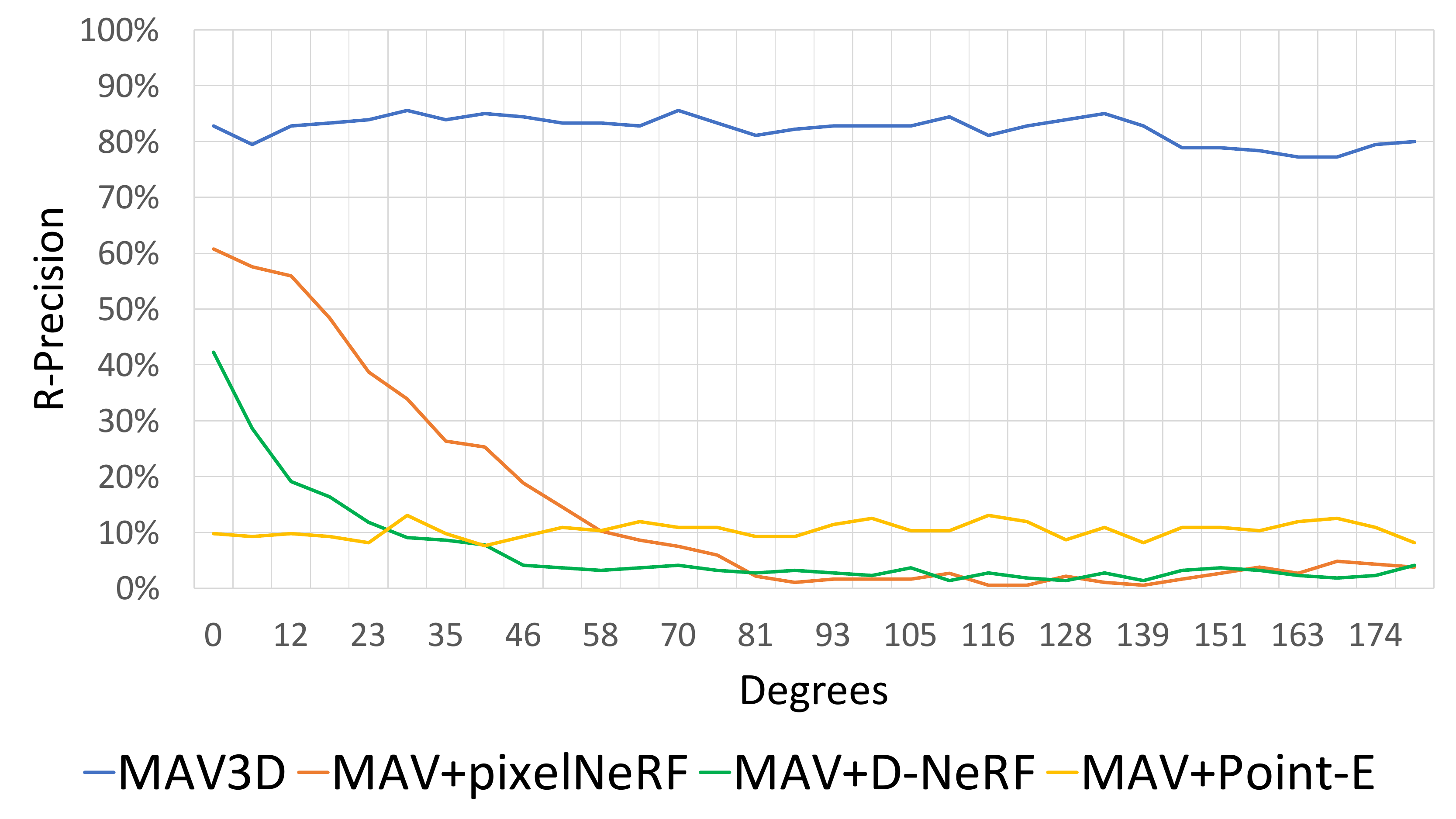}\vspace{-2mm}
\caption{
  R-Precision per viewing angle.
  We render per method from cameras in a circle around the scene.
}%
\label{fig:clip_by_angle}
\vspace{-2mm}
\end{figure}

\textbf{Metrics.}
We evaluate the generated videos using
CLIP R-Precision~\cite{jain2022zero}, which measures the consistency between the text and the generated scene.
The reported metric is the retrieval accuracy of input prompts from rendered frames.
We use the \texttt{ViT-B/32} variant of CLIP~\cite{wang2022clip} and extract frames in varying views and time steps.
We also use four qualitative metrics by asking human raters their preferences among two generated videos based on:
(i) video quality;
(ii) faithfulness to the textual prompt;
(iii) amount of motion;
and (iv) realism of motion.
We evaluated all baselines and ablations on the text prompts splits which were used in~\cite{singer2022make}.

\textbf{Samples.}
We show samples in Figures~\ref{fig:teaser} and \ref{fig:scene_sketching_comparisons}.
For more detailed visualization, please see \github.

\subsection{Results}
\textbf{Text-to-4D comparison.}
As there were no previous methods for Text-To-4D, we established three baselines for comparison.
The baselines are based on a T2V generative method~(\mav~\cite{singer2022make}), which generates a sequence of 2D frames from a text prompt.
Once generated, the sequence of 2D frames is transformed into a sequence of 3D scene representations using three different methods. The first sequence is produced by applying a one-shot neural scene renderer (Point-E~\cite{nichol2022point}, the second via pixelNeRF~\cite{yu2021pixelnerf}) on each frame independently, and third by applying D-NeRF~\cite{pumarola2021d} combined with camera position extracted using COLMAP~\cite{colmap}. We denote these adapted baselines as MAV+Point-E, MAV+pixelNeRF and MAV+D-NeRF, respectively.
For the Point-E baseline, we use the \texttt{base1B} variant released by the authors.
For the D-NeRF baseline, we use only videos with valid COLMAP results and substantial camera motion. Results are presented at the top of Table~\ref{table:main_exp}.
As can be seen, our method surpasses the naive baselines in the objective R-Precision metric and is highly preferred by human raters across all metrics.


\setlength{\tabcolsep}{2pt}
\begin{table}
  \centering
  \caption{Comparison with baselines~(R-Precision and human preference). \textbf{Human evaluation is shown as a percentage of majority votes in favor the baseline compared to our model in the specific setting}.}
  \label{table:main_exp}
  \scalebox{0.85}{
  \begin{tabular}{c|lc|cccc}
  \toprule\noalign{\smallskip}
  D & Model & R-  & Video & Text- & More & Realistic \\
   & & Precision$\uparrow$ & quality & align. & motion  & motion \\
  \noalign{\smallskip}
  \hline
  \noalign{\smallskip}
   \multirow{ 4}{*}{4D} & MAV+pixelNeRF& 8.8  & 0.11  & 0.16 & 0.18 & 0.16\\
  & MAV+Point-E & 10.6 & 0.20 & 0.06 & 0.26 & 0.17 \\
  & MAV+D-NeRF & 6.2 & 0.17 & 0.18 & 0.19 & 0.30 \\
  & \textbf{\fd} & 83.7  & -  & - & - & -\\
  \hline
  \noalign{\smallskip}
   \multirow{3}{*}{3D} & Stable DF  & 66.1 & 0.46 & 0.36 & - & -\\
  & Point-E & 12.6 & 0.15 & 0.16 & - &  - \\
  & \textbf{MAV3D\textbackslash t} & 82.4 & - & - & - &  - \\
  \hline
  \noalign{\smallskip}
   \multirow{ 2}{*}{Video} & MAV  & 86.6  & 0.73 & 0.63 & 0.64 & 0.62 \\
   & \textbf{MAV3D\textbackslash z} & 79.2 & -  & - & - & -\\
  \bottomrule
  \end{tabular}}
  \vspace{-4mm}
\end{table}

Furthermore, we explore our method performance for different camera viewing angles. This is done by calculating R-precision for frames rendered from different camera positions. Specifically, given a camera position $d=(R,\theta,\phi)$, we fix the zoom, $R$, and the tilt, $\theta$, and report R-Precision for different pan values - $\phi$. In Fig~\ref{fig:clip_by_angle} we show our method is able to render the scene consistently across viewing angles while the MAV+D-NeRF and MAV+pixelNeRF performance deteriorates as $\phi$ increases. MAV+Point-E is also able to maintain a consistent R-Precision score.

\textbf{Text-to-3D comparison.}
To evaluate our method on the Text-to-3D~(3D) task, we remove the temporal dimension from our rendered video. Specifically, we sample \fd from a single time step, and denote this reduction as \textbf{MAV3D\textbackslash t}.
We compare this variant with: (i) Stable-DreamFusion~(Stable-DF)~\cite{stabledreamfusion}, a public re-implementation of DreamFusion~\cite{poole2022dreamfusion}, and, (ii) Point-E~\cite{nichol2022point}, which generates a point cloud given a text prompt.
The results are presented in Table~\ref{table:main_exp}. In this setting, the input to each baseline is the text prompt. Since the \texttt{base1B} variant of Point-E expects image input, we utilize a T2I model that generates an image which is then fed to the model.
As can be seen, our model is preferred over these variants in quality and text alignment.




\textbf{Text-to-Video comparison.}
To evaluate our method on the sub-task of T2V~(Video), we remove the depth dimension from our method output.
Concretely, we sample frames from our model on specific viewing directions~(front, back, and side),
reduceing our method from temporal dynamic scene generation to video generation.
We denote this reduction as \textbf{MAV3D\textbackslash z}.
This variant is compared to videos generated with \mav by appending the viewing direction to the textual input prompt.
Results are presented at Table~\ref{table:main_exp}. 
Note that our method utilizes \mav as a training objective and is thus bounded by its performance.

\setlength{\tabcolsep}{1pt}
\begin{table}
\centering
\caption{Ablation study (R-Precision and human preference ($\%$) . \textbf{Human evaluation is shown as a percentage of majority votes in favor the baseline vs our model}. 
}
\label{table:abl_study}
\scalebox{0.95}{
\begin{tabular}{lc|cccc}
\toprule\noalign{\smallskip}
Model & R- & Video & Text- & More & Realistic \\
& Precision$\uparrow$ & quality & align. & motion  & motion \\
\noalign{\smallskip}
\hline
\noalign{\smallskip}
 \small{w/o SR} & 84.3 & 0.41 & 0.34 & 0.42 & 0.36 \\
\small{w/o pretraining} & 63.5 & 0.27 & 0.44 & 0.46 & 0.35\\
\small{w/o dynamic camera} & 83.6 & 0.54 & 0.48 & 0.35 & 0.42 \\
\small{w/o gaussian anneal.}  & 76.3 & 0.47 & 0.50 & 0.45 & 0.38 \\
\small{with D-NeRF} & 81.9 & 0.45 & 0.47 & 0.50 & 0.47 \\
\small{with Instant NGP} & 78.4 & 0.36 & 0.40 & 0.48 & 0.42 \\

\bottomrule
\end{tabular}}
\vspace{-4mm}
\end{table}

\subsection{Ablation study}
\label{sec:ablation_sec}
An ablation study of human preference and R-precision is provided in Tab.~\ref{table:abl_study} to assess the effectiveness of our different contributions.

 \textbf{Ablation on different training stages.} (i) \textit{without SR:} a model trained without the scene super-resolution fine-tuning, for the same number of steps as \fd (stage 3). As can be seen, human raters favors the model trained with SR in both quality, text alignment and motion. In addition, as demonstrated in Fig.~\ref{fig:SR_ablat_fig}, the super-resolution fine-tuning enhances the quality of the rendered videos, resulting in high-resolution videos with finer details (e.g. the hands for the fox) and less noise. (ii) \textit{without pre-training:} As illustrated in Tab.~\ref{table:abl_study} and in Fig.~\ref{fig:static_pretrain_fig}, directly optimizing the dynamic scene (without the static scene pre-training) for the same number of steps as \fd, results in much lower scene quality or poor convergence:
 the model trained with static-pretraining is preferred for video quality and realistic motion in 73\% and 65\% of cases, respectively. Additional ablation of the number of static pretrain steps is available in the supplementary.

\begin{figure*}[t!]
    \setlength{\tabcolsep}{1.5pt}
    {\small
    \begin{tabular}{c c }
        \vspace{-0.1cm}
        \includegraphics[width=0.50\textwidth]{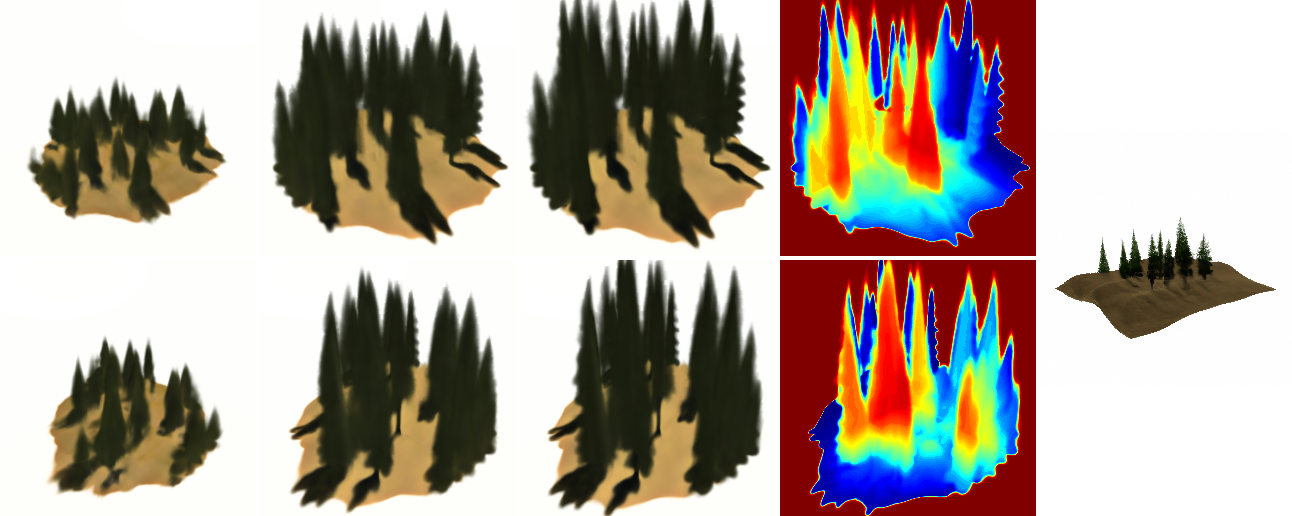} &
        \hspace{0.1cm}
        \includegraphics[width=0.50\textwidth]{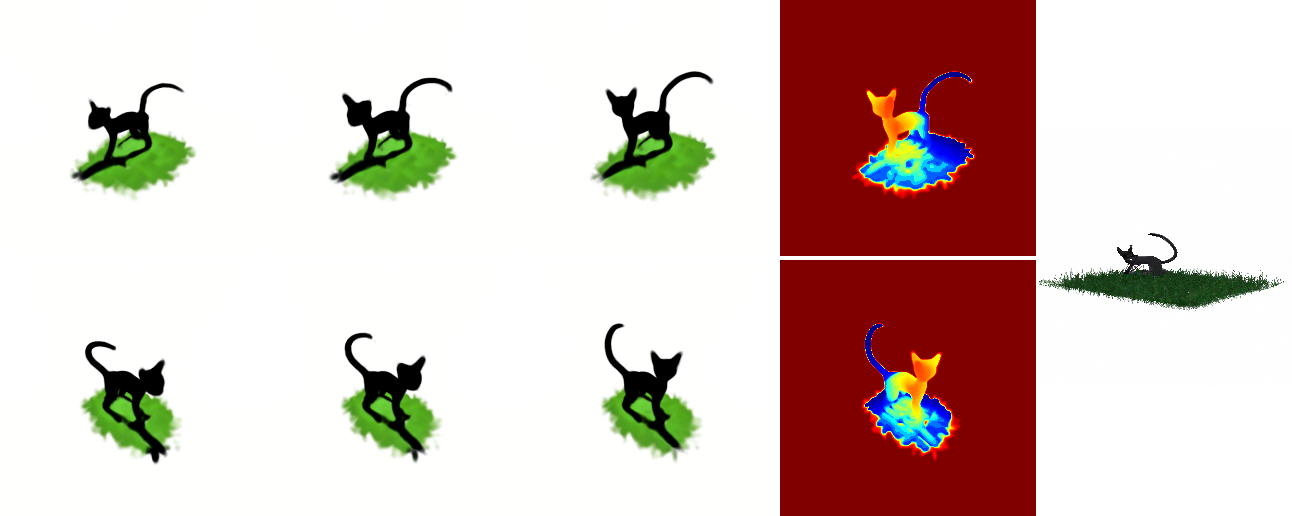} \\
        \qquad\qquad\quad\quad 3D Video \quad\qquad\qquad Depth Image \hskip0.5em\relax Condition &
        \qquad\qquad\quad\quad 3D Video \qquad\qquad\qquad Depth Image \hskip0.5em\relax Condition \\
        \vspace{-0.1cm}
        \includegraphics[width=0.50\textwidth]{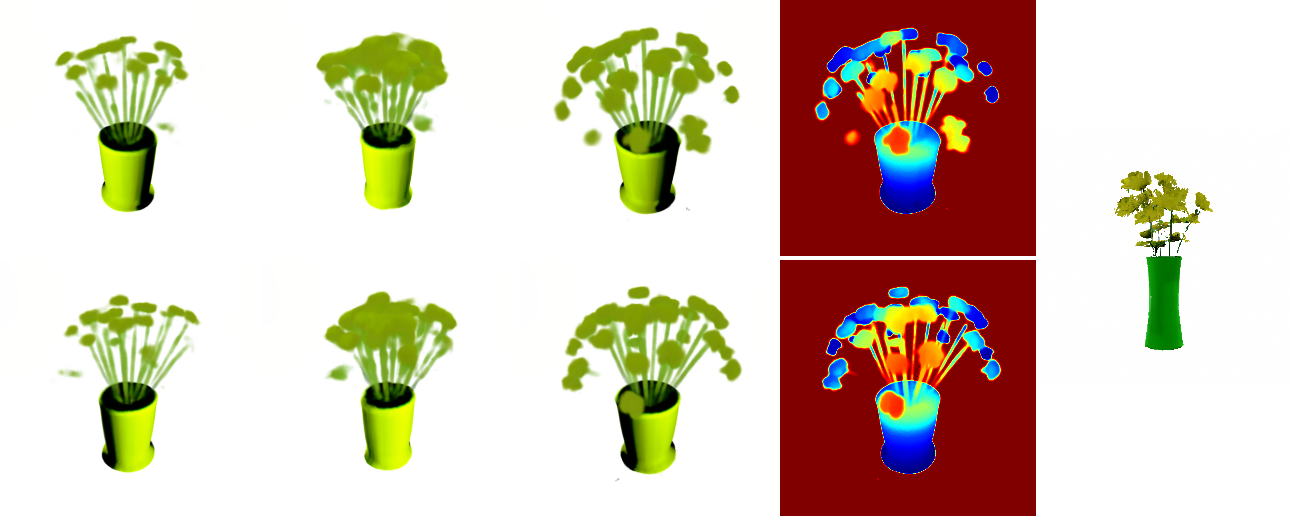} &
        \hspace{0.1cm}
        \includegraphics[width=0.50\textwidth]{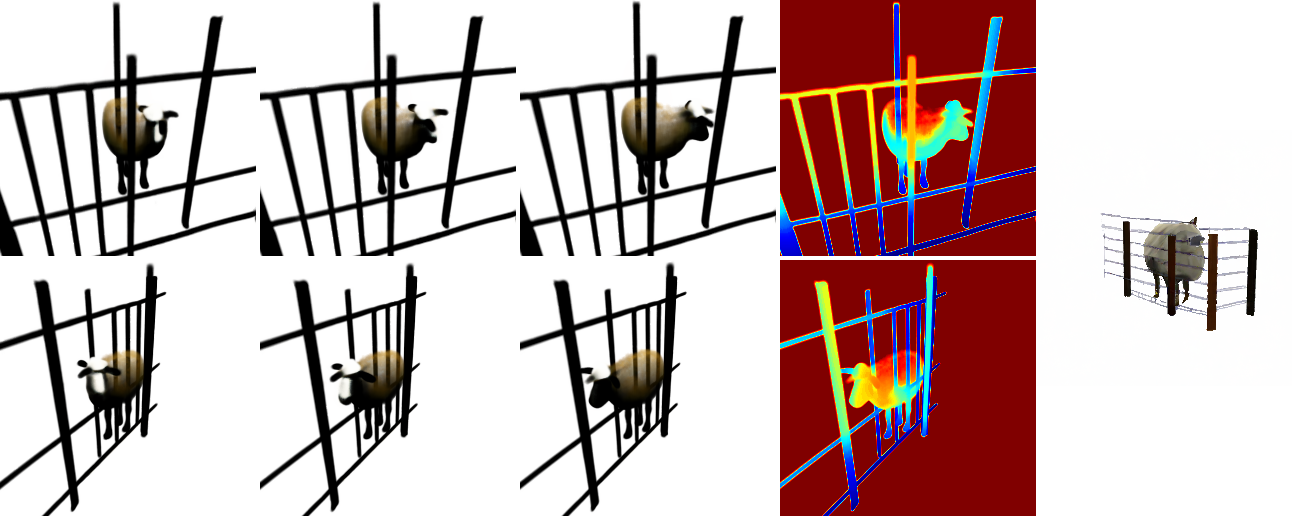} \\
        \qquad\qquad\quad\quad 3D Video \quad\qquad\qquad Depth Image \hskip0.5em\relax Condition &
        \qquad\qquad\quad\quad 3D Video \qquad\qquad\qquad Depth Image \hskip0.5em\relax Condition
    \end{tabular}
    }
    \vspace{-0.2cm}
    \caption{\textbf{Image to 4D application.} Given an input image 
    we extract its CLIP embedding, and use it to condition \fd.} 
    \vspace{-0.1cm}
    \label{fig:image_to_4d}
\end{figure*}

\textbf{Ablation on motion regularizers components.} (i) \textit{dynamic camera:} here, we train a variant of our method in which the camera position is fixed across all frames. We observe that videos rendered using this variant obtain less motion, and suffer from multi-face object (see Fig.~\ref{fig:dynamic_camera_fig}). This may explain the relatively high R-Precision score (since with multiple faces, the object can be recognized more easily).
(ii) \textit{Gaussian annealing:} Extending the spatial bias of the model (to focus on the larger "blob") leads to renderings with larger and more realistic motion.

\textbf{Ablation on NeRF backbone.} (i) \textit{with D-NeRF:} to quantify the contribution of the temporal NeRF, we analyzed a variant of our method in which we replaced our temporal NeRF backbone (HexPlane) with D-NeRF~\cite{pumarola2021d, fang2022fast}. While HexPlane is slightly preferred in terms of overall quality and realistic motion, our approach is not sensitive to the dynamic backbone, demonstrating the robustness of our method. 
(ii) \textit{with Instant-NGP:} here we replace our static NeRF backbone with Instant-NGP~\cite{mueller2022instant}. This variation employs hash encoding and includes a color network that emits radiance values, and it is significantly less preferred.



\subsection{Real-time rendering}

Applications such as virtual reality and games that use traditional graphic engines require assets in a standard format such as textured meshes.
The HexPlane model can be easily converted to animated meshes as follows.
First, the marching cube algorithm is used to extract a simplicial mesh from the opacity field generated at each time $t$, followed by mesh decimation (for efficiency) and removal of small noisy connected components.
The XATLAS~\cite{xatlas} algorithm is used to map the mesh vertices to a texture atlas and the texture is initialized using the HexPlane colors averaged in small spheres centred around each vertex.
Finally, the texture is further optimized to better match a number of example frames rendered by HexPlane using a differentiable mesh rendered.
This results in a collection of texture meshes that can be easily played back in any off-the-shelf 3D engine~(see \github for running examples).

\subsection{Image To 4D}
Given an input image, we can use it to condition our method to generate its 4D asset. 
Similar to~\cite{singer2022make}, instead of conditioning the T2I and T2V components on the output of the prior, we condition them directly on the CLIP~\cite{yu2022scaling} embedding of the input image. This allows us to create a 4D asset that shares the same semantics as the input image. For this experiment we took images provided by \cite{nichol2022point} that were used there for the Image-to-3D task. Fig.~\ref{fig:image_to_4d} and Fig.~\ref{fig:image_to_4d_supp} demonstrate our method ability to generate both depth and motion from a given input image, resulting in a 4D asset.


\label{experiments}

\section{Discussion}
\label{discussion}


Creating animated 3D content is challenging as the tools available today are manual and catered to professionals. While current models can generate static 3D objects, synthesizing dynamic scenes is more complex. Unlike images and videos, where large amounts of captioned data are readily available, 4D models are scarce, with or without text descriptions.

In this work, we present \fd, a new approach that employs several diffusion models and dynamic NeRFs to integrate world knowledge into 3D temporal representations. \fd expands the functionality of previously established diffusion-based models, enabling them to generate dynamic scenes as described in text from a variety of viewpoints. 

While our model is a step towards zero-shot temporal 3D generation, it also has several limitations. The conversion of dynamic NeRFs to a sequence of disjoint meshes for real-time applications is inefficient and could be improved if trajectories of vertices were predicted directly. Also, utilizing super-resolution information has improved the quality of the representation, but further improvement is needed for higher-detailed textures. Finally, the quality of the representation is dependent on the ability of the T2V model to generate videos from various views. While utilizing view-dependent prompts helps mitigating the multi-face problem, further control to the video generator would be beneficial.


\ifdefined\arxivversion 
\section{Acknowledgements}
\label{ackz}

We would like to thank Vincent Cheung, Natalia Neverova, Kelly Freed, Oran Gafni, Thomas Hayes, and Vikram Voleti for their contributions to this project.

\fi


\bibliography{MAV3D}
\bibliographystyle{icml2023}


\newpage
\appendix
\onecolumn
\section{Appendix}

\subsection{Ablation on static scene pretrainig steps.} 

\begin{figure*}[ht!]
\label{num_steps_astatic}
    \centering
    \includegraphics[
    width=0.7\textwidth]{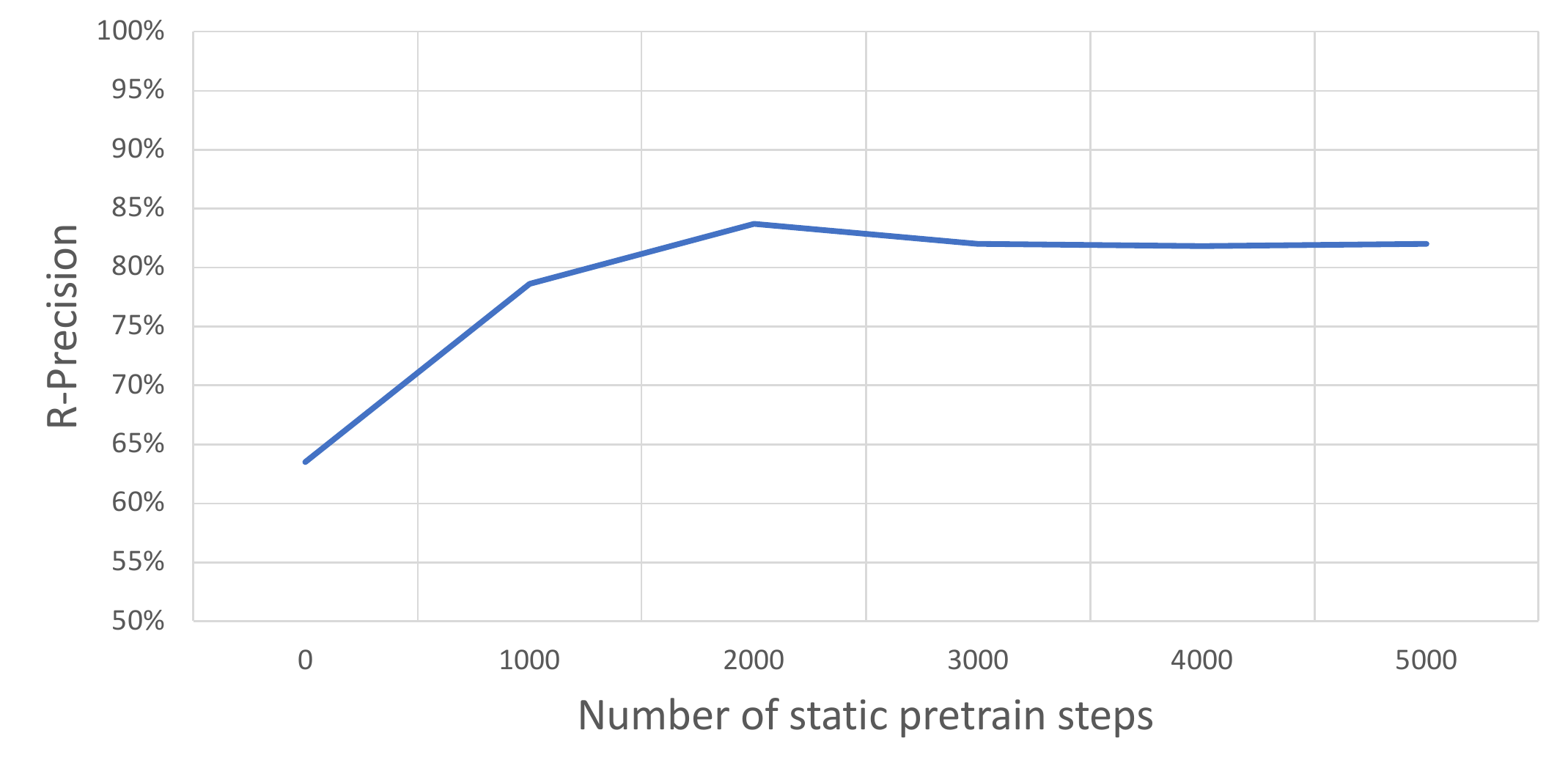}
    \caption{\textbf{Ablation on the number of static pretraining steps}}
    \label{fig:num_steps_ablation}
    \vspace{-3mm}
\end{figure*}

In Fig.~\ref{fig:num_steps_ablation}, we analyze the different number of steps required for the static scene pretraining. As observed, directly optimizing the dynamic scene leads to sub-optimal convergence (R-Precision of 63.5\%). Furthermore, $2000$ iterations are sufficient for achieving high-quality results (R-Precision of $83.7\%$). An interesting observation is that further increasing the number of static pretraining steps beyond $2000$ does not improve the quality of the scene representation. 

\subsection{Implementation details}
\label{implement_details}
\paragraph{Architecture details.} We adopt a multi-resolution grid encoding architecture, similar to ~\cite{mueller2022instant} with $7$ levels of resolutions, spanning from a minimum resolution of $16 \times 16$ up to a maximum resolution of $2048 \times 2048$. We use $5$ layers MLP with $128$ hidden units, each followed by ReLU activation.
Similar to ~\cite{poole2022dreamfusion}, we train another $3$ layers MLP network with $64$ hidden units for the background representation. The background is encoded using frequency encoder and is not conditioned on the time.

\begin{figure*}[ht!]
    \centering
    \setlength{\tabcolsep}{1.5pt}
    {\small
    \begin{tabular}{c c }
        \vspace{-0.1cm}\includegraphics[width=0.49\textwidth]{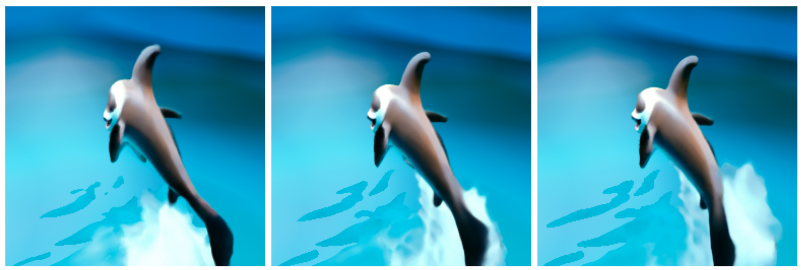} &
        \hspace{0.1cm}
        \includegraphics[width=0.49\textwidth]{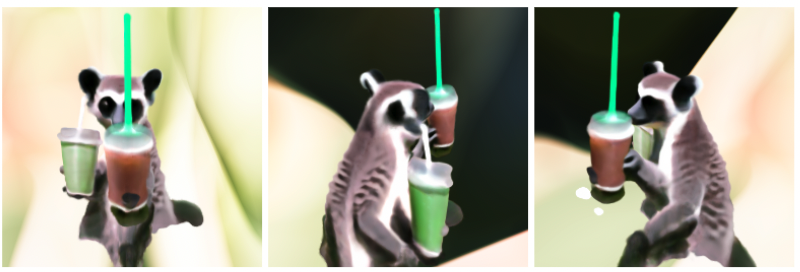} \\
        \vspace{-0.1cm}\includegraphics[width=0.49\textwidth]{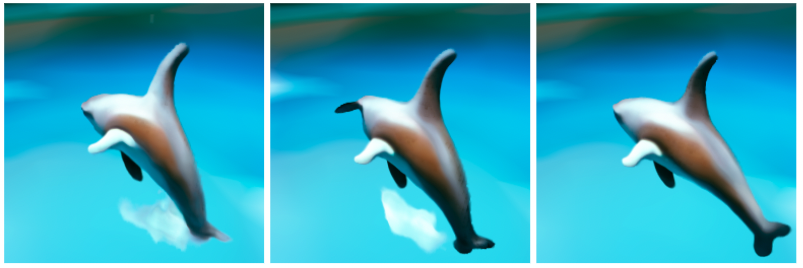} &
        \hspace{0.1cm}
        \includegraphics[width=0.49\textwidth]{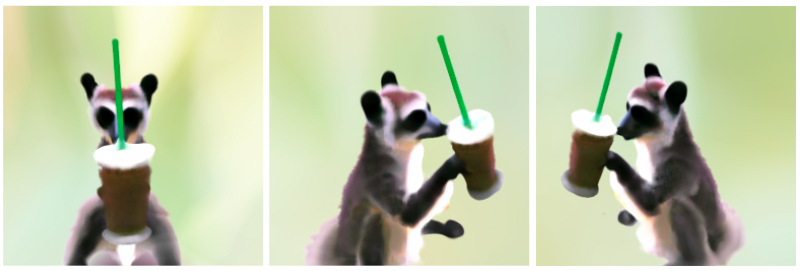} \\
        \vspace{0.2cm} A dolphin jumping out of the water & A lemur drinking boba \\
    \end{tabular}
    }
    \vspace{-0.2cm}
    \caption{An ablation on the \textbf{dynamic camera} component. \textbf{Top:} without dynamic camera. \textbf{Bottom:} MAV3D results. The figures on the left demonstrate training with dynamic camera results in larger and more complicated motion (The dolphin manages to jump out of the water). The figures on the right demonstrate how the dynamic camera mitigates the multi-face problem.
    } 
    \vspace{-0.1cm}
    \label{fig:dynamic_camera_fig}
\end{figure*}

\begin{figure*}[ht!]
    \centering
    \setlength{\tabcolsep}{1.5pt}
    {\small
    \begin{tabular}{c c }
        \vspace{-0.1cm}\includegraphics[width=0.49\textwidth]{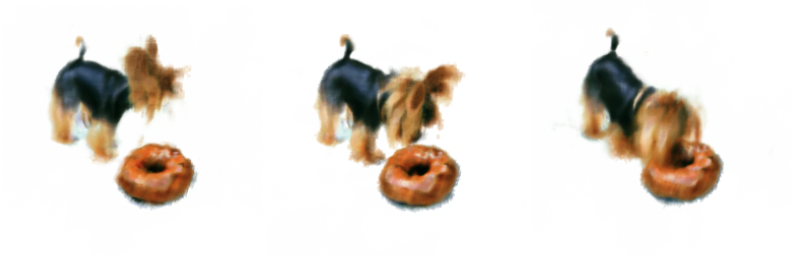} &
        \hspace{0.1cm}
        \includegraphics[width=0.49\textwidth]{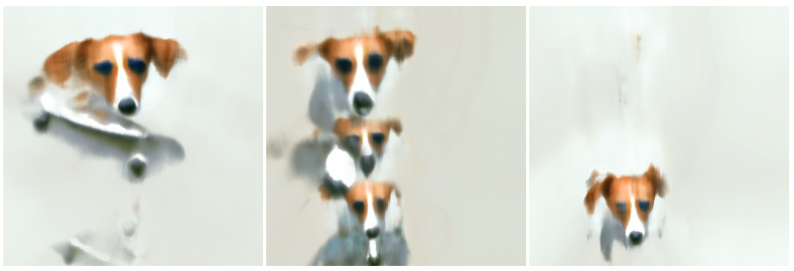} \\
        \vspace{-0.1cm}\includegraphics[width=0.49\textwidth]{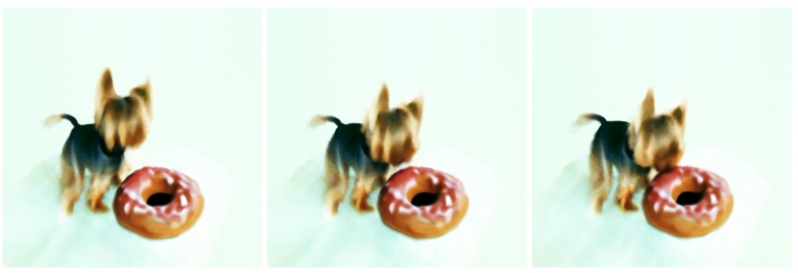} &
        \hspace{0.1cm}
        \includegraphics[width=0.49\textwidth]{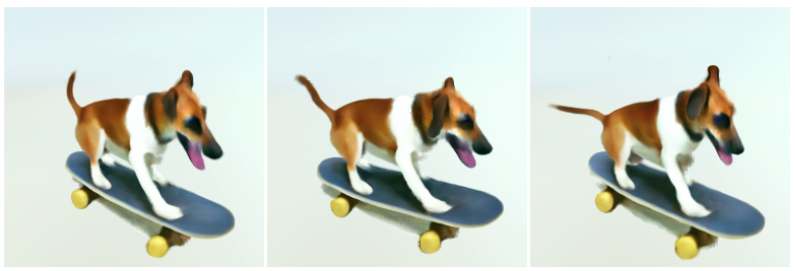} \\
        \vspace{0.2cm} A yorkie dog eating a donut & A dog riding a skateboard \\
    \end{tabular}
    }
    \vspace{-0.2cm}
    \caption{An ablation on \textbf{static scene pre-training}. \textbf{Top:} without static scene pretraining. \textbf{Bottom:} MAV3D results. As can be seen in the figures on the left, pretraining the model on static scene leads to higher quality renderings. As can be seen in the figures on the right, training directly the dynamic scene (both the 3D representation and the temporal dimension) may result in lack of convergence.
    } 
    \vspace{-0.1cm}
    \label{fig:static_pretrain_fig}
\end{figure*}

\subsection{Training details}
\label{training_details}
Unless otherwise noted, we use a batch size of $8$ and sample $128$ points along each ray. During training, the camera position is randomly sampled in spherical coordinates, with radius in range $[1,1.5]$, and the scene is bounded in box with radius $1$. 


\paragraph{Dynamic Camera.}
 As the utilized T2V model generates videos with a moving camera $\{C_t\}_{t=1}^T$, we propose to bridge the distribution gap between the generated videos (by \mav) and the rendered videos (by \nerf)  using a dynamic camera position. Training the model using dynamic camera trajectory simulates the movement of a real moving camera and thus enhancing the realism and coherence of the motion learned by the model. \\
To this end, given the first camera position $C_1=(R,\theta,\phi)$ (randomly sampled in spherical coordinates bounded to  the range $R \in [1,1.5]$, $\theta \in [0, \frac{2\pi}{3}]$, and $\phi \in [0,2\pi]$), we employ a random camera trajectory $\{C_1 + (i-1) \cdot \mathbf{d}\}$, in which camera $C_i$ is displaced by $(i-1) \cdot \mathbf{d}$ at each time-point $i$. Specifically, $\mathbf{d} = (\Delta R,  \Delta \theta, \Delta \phi)$, where
$\Delta R \sim \mathcal{U}[\frac{1-R}{F}, \frac{1.5-R}{F}]$, 
$\Delta \theta \sim \mathcal{U}[-\frac{\pi}{4F}, \frac{\pi}{4F}]$, 
$\Delta \phi \sim \mathcal{U}[-\frac{\pi}{2F}, \frac{\pi}{2F}]$.

As demonstrated in the experiments, a dynamic camera also helps reducing the amount of temporal artifacts such as unrealistic number of object parts, as it has visual of the object from multiple directions in the same sample.

\paragraph{Gaussian annealing.}
When optimizing a static scene model, incorporating a spatial bias towards the center of the scene can be beneficial, as it helps focusing in the center of the scene rather than directly next to the sampled cameras~\cite{poole2022dreamfusion}. The added noise, which is parameterized using a Gaussian PDF, causes a small “blob” of density to the origin of the scene. 
However, in dynamic scene rendering, objects that were originally centered at the origin may move to surrounding areas, making this bias less effective. Enlarging the standard deviation of the bias added to the density $\tau$ can encourage density not only in the center of the scene, but also in nearby locations, further enhancing the realism of the motion:  
\begin{align}
\begin{split}
\lambda_\tau \cdot \exp\Big(-\frac{\lVert\mu\rVert^2}{2\sigma(ts)^2_\tau}\Big)
\end{split}
\end{align}
Where $\sigma$ is a function of the training step, $ts$. In order to anneal the bias for $M=5000$ training steps from a minimum value $\sigma_{min}=0.2$ to a maximum value $\sigma_{max}=2.0$, we define a linear function as follows: $\sigma(ts)=min(\sigma_{max}, \sigma_{min}+(\sigma_{max}-\sigma_{min}) \cdot\frac{ts}{M})$.

\myparagraph{Optimization.} We train the model using the Adam optimizer, with cosine decay scheduler, starting from learning rate of 1e-3. 
The static scene representation is trained on rendered images of $64 \times 64$ for 5000 iterations with a total runtime of around 30 minutes. The dynamic stage is trained on rendered videos of $64 \times 64 \times 16$ for 5,000 iterations with a total runtime of around 3 hours. Lastly, the super resolution phase is trained on rendered videos of $256 \times 256 \times 16$ for another 2000 iterations with a total runtime of 3 hours. All runtimes were measured on 8 NVIDIA A100 GPUs.
During inference, by leveraging the continuous time range, we render videos of $256 \times 256 \times 64$.

\myparagraph{Training objective}
In order to encourage the model to make harder predictions if a specific pixel is an object or background, we add the following soft binary cross entropy regularization:

$$
- \sum_{i,j} \Tau_{i,j} log(\Tau_{i,j}) + (1-\Tau_{i,j}) log(1-\Tau_{i,j})
$$

where $\Tau_{i,j}$ denotes the accumulated density along the ray of pixel $(i,j)$, (i.e., the probability that the entire ray does not hit any particle). This regularization is added to the loss with a weight of $10^{-3}$.

The weight of the variational loss described in Sec.~\ref{sec:totalvarloss} is $10^{-3}$.

\subsection{Additional results for Image-to-4D}

\begin{figure*}[h!]
    \setlength{\tabcolsep}{1.5pt}
    {\small
    \begin{tabular}{c c }
        \vspace{-0.1cm}
        \includegraphics[width=0.49\textwidth]{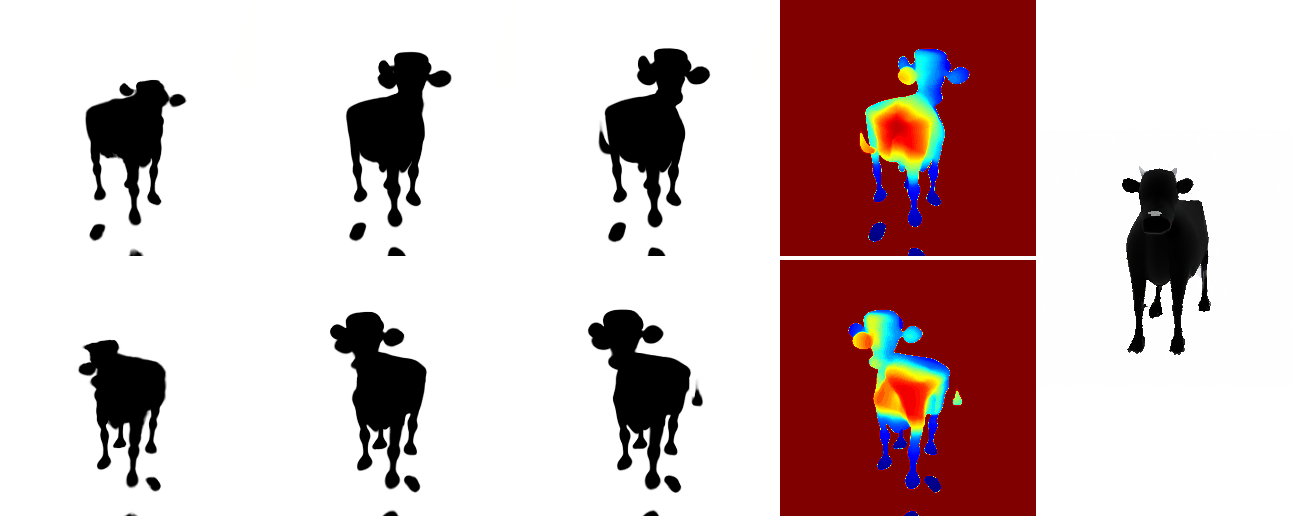} &
        \hspace{0.1cm}
        \includegraphics[width=0.49\textwidth]{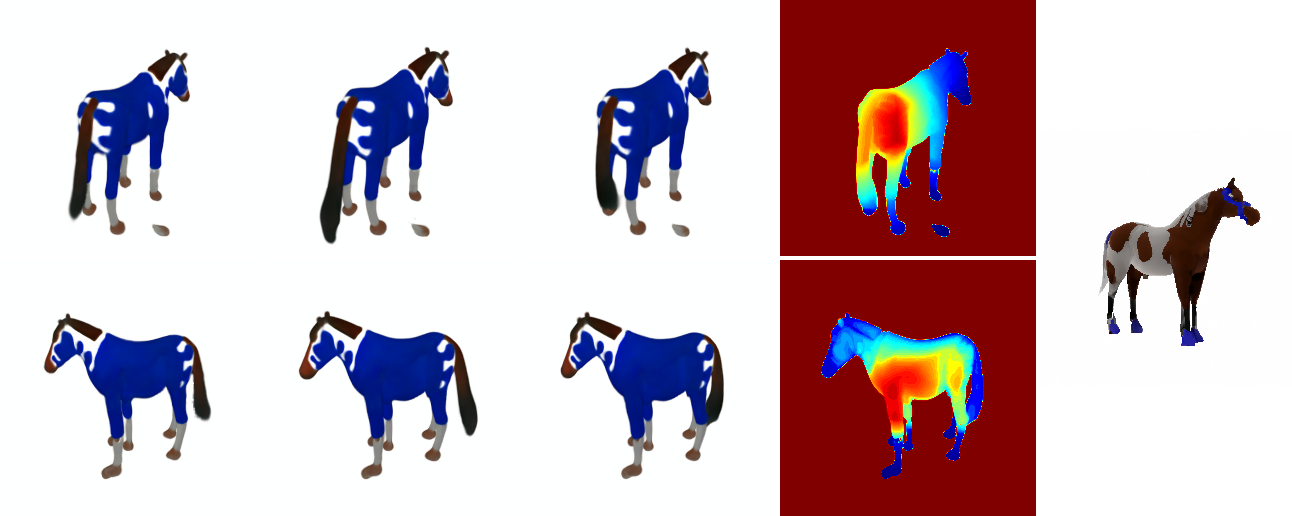} \\
        \qquad\qquad\quad\quad 3D Video \quad\qquad\qquad Depth Image \hskip0.5em\relax Condition &
        \qquad\qquad\quad\quad 3D Video \qquad\qquad\qquad Depth Image \hskip0.5em\relax Condition \\
        \vspace{-0.1cm}
        \includegraphics[width=0.49\textwidth]{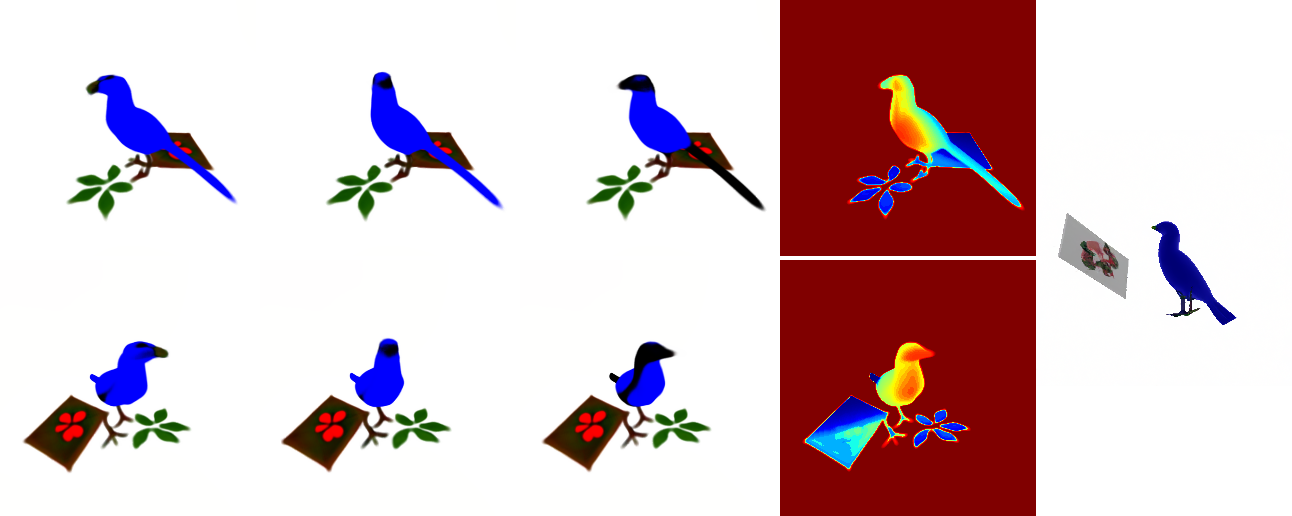} &
        \hspace{0.1cm}
        \includegraphics[width=0.49\textwidth]{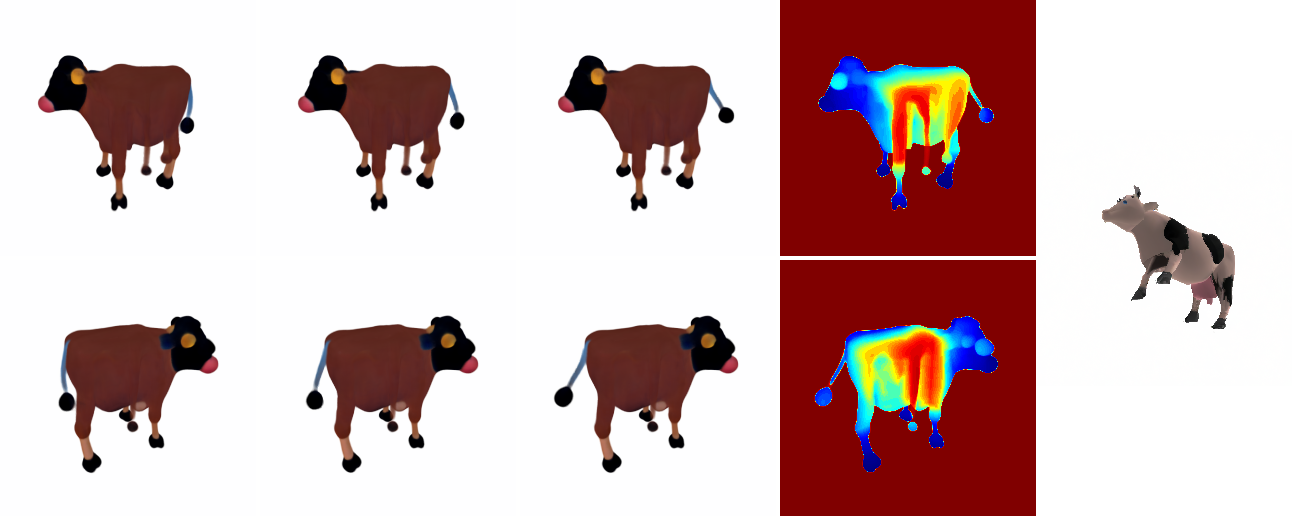} \\
        \qquad\qquad\quad\quad 3D Video \quad\qquad\qquad Depth Image \hskip0.5em\relax Condition &
        \qquad\qquad\quad\quad 3D Video \qquad\qquad\qquad Depth Image \hskip0.5em\relax Condition \\
        \vspace{-0.1cm}
        \includegraphics[width=0.49\textwidth]{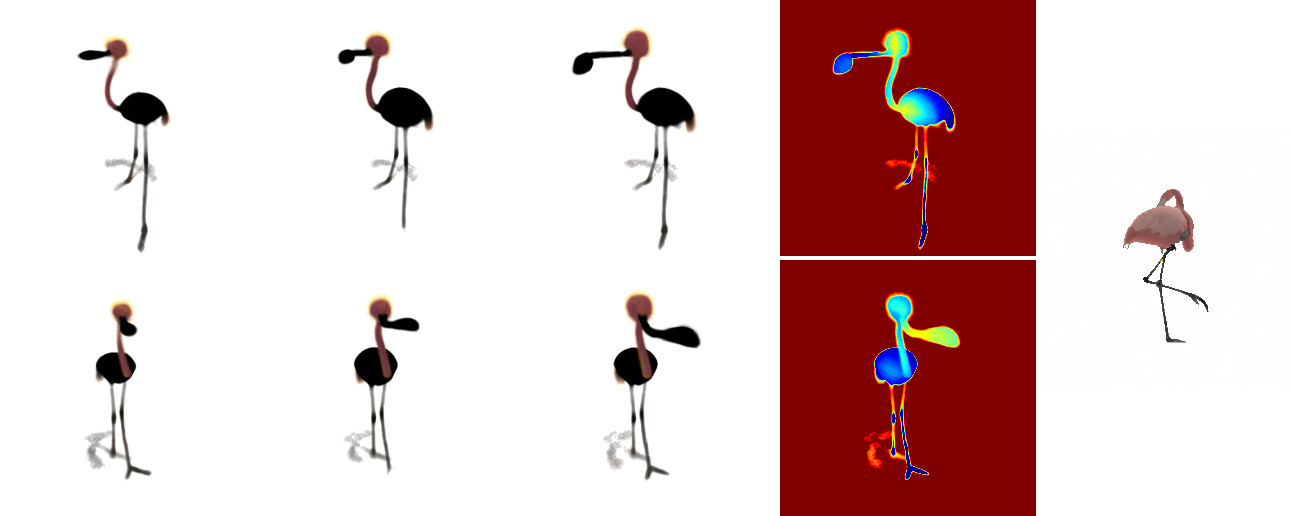} &
        \hspace{0.1cm}
        \includegraphics[width=0.49\textwidth]{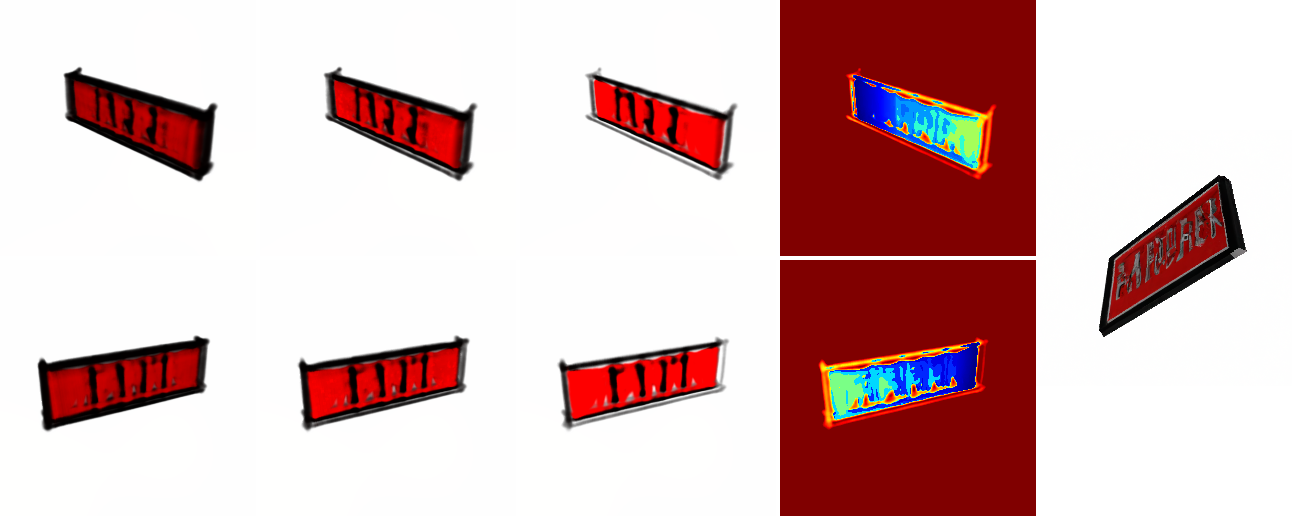} \\
        \qquad\qquad\quad\quad 3D Video \quad\qquad\qquad Depth Image \hskip0.5em\relax Condition &
        \qquad\qquad\quad\quad 3D Video \qquad\qquad\qquad Depth Image \hskip0.5em\relax Condition
    \end{tabular}
    }
    \vspace{-0.2cm}
    \caption{Additional results for the Image to 4D application.} 
    \vspace{-0.1cm}
    \label{fig:image_to_4d_supp}
\end{figure*}



\end{document}